%% file: iclr2017_workshop.tex
\documentclass{article} 
\usepackage{iclr2017_workshop,times}
\usepackage{hyperref}
\usepackage{url}

\usepackage{graphicx}
\usepackage{amsmath}
\usepackage{booktabs}
\usepackage{tabu}
\usepackage[tracking=true]{microtype}
\usepackage{wrapfig}

\title{Natural Language Generation in \linebreak Dialogue using Lexicalized and \linebreak Delexicalized Data}

\author{
	Shikhar Sharma \\
	Microsoft Maluuba
	\And
	Jing He\thanks{This work was done while Jing He was at Maluuba (now Microsoft Maluuba)} \\
	AdeptMind
	\And
	Kaheer Suleman \\
	Microsoft Maluuba
	\AND
	Hannes Schulz \\
	Microsoft Maluuba
	\And
	Philip Bachman \\
	Microsoft Maluuba
}

\graphicspath{{images/}{./}}
\begin{document}

\maketitle

\input{abstract.tex}
\input{introduction.tex}

\input{model.tex}
\input{evaluation.tex}
\input{results.tex}
\input{conclusion.tex}

\bibliography{iclr2017_workshop}
\bibliographystyle{iclr2017_workshop}

\newpage
\appendix

\section{Datasets and human evaluation}
\label{sec:experiments}
\begin{wraptable}{l}{75mm}
	\small
	\vspace{-8mm}
	\caption{\label{table:dataset-stats} Statistics for the CF and DSTC2 datasets}
	\vspace{2mm}
	\centering
	\begin{tabu}to0.52\columnwidth{@{}X[l,1.5]*2{X[r]}X[c,0.10]*2{X[r]}@{}}
		\toprule
		& \multicolumn{2}{c}{\bf CF} & & \multicolumn{2}{c}{\bf DSTC2} \\ \cmidrule{2-3} \cmidrule{5-6}
		\bf             & \bf Train & \bf Test       & & \bf Train & \bf Test          \\ \midrule
		\#words         & 15,143    & 2,033          & & 240,337   & 127,858           \\
		\#sentences     & 1,200     & 211            & & 15,611    & 9,890             \\
		\#vocabulary    & 690       & 286            & & 660       & 166               \\
		\bottomrule
	\end{tabu}
	\vspace*{-2mm}
\end{wraptable}

\input{dataset.tex}

\paragraph{Human evaluation of responses}
\label{subsec:humaneval}
We selected a random set of $100$ dialogue acts from each dataset's test set and the
corresponding top response generated by all of the models, then asked $5$ human judges to
score them on a scale of $1$ to $5$, with $1$ indicating least appropriate for the given
dialogue acts and $5$ indicating most appropriate. In each trial, we presented $4$
sentences to the judges, each from a different model, along with the corresponding
dialogue acts. The judges were informed that all sentences had been generated from
different models and not presented in any particular order.

\section{Training details}
\paragraph{Hyper-parameters}
The number of layers in the decoder, the decoder hidden state dimension, the
encoder hidden state dimension and the word embedding dimensionality
 are set
using the validation set. The reading coefficient of the sc-LSTM units,
$\alpha$, is set to $1$ and the maximum length of $T$ is set to $30$.
 We employ the
Adam optimizer~\citep{DBLP:journals/corr/KingmaB14} for training and
 we apply a dropout
\citep{DBLP:journals/jmlr/SrivastavaHKSS14} of $0.5$ at all non-recurrent connections.

\end{document}

%% file: abstract.tex
\vspace*{-5mm}

\begin{abstract}
\label{sec:abstract}
Natural language generation plays a critical role in spoken dialogue systems. We
present a new approach to natural language generation for task-oriented dialogue using recurrent neural networks
in an encoder-decoder framework. In contrast to previous work, our model uses both
lexicalized and delexicalized components i.e. slot-value pairs for dialogue acts,
with slots and corresponding values aligned together. This
allows our model to learn from all available data including the slot-value pairing, rather than being restricted to
delexicalized slots. We show that this helps our model
generate more natural sentences with better grammar. We further improve our model's
performance by transferring weights learnt from a pretrained sentence auto-encoder. Human
evaluation of our best-performing model indicates that it generates sentences which
users find more appealing.
\end{abstract}

%% file: introduction.tex
\section{Introduction}
\label{sec:introduction}

Traditionally, task-oriented spoken dialogue systems (SDS) rely on template-based,
hand-crafted rules for natural language generation (NLG). However, this approach does
not scale well to complex domains and datasets. Previous papers have explored
alternatives using corpus-based \mbox{n-gram} models~\citep{DBLP:journals/csl/OhR02}, tree-based
models~\citep{DBLP:journals/jair/WalkerSMP07}, SVM
rerankers~\citep{DBLP:conf/acl/KondadadiHS13}, and Reinforcement Learning
models~\citep{DBLP:conf/eacl/RieserL10}.

Recently, models based on recurrent neural networks (RNNs) have been successfully
applied to natural language generation tasks such as image
captioning~\citep{DBLP:conf/icml/XuBKCCSZB15,DBLP:conf/cvpr/KarpathyL15}, video
description~\citep{DBLP:conf/iccv/YaoTCBPLC15,thesis/Sharma16}, and machine
translation~\citep{DBLP:journals/corr/BahdanauCB14}. In the domain of task-oriented SDS,
RNN-based models have been used for NLG in both traditional multi-component processing
pipelines~\citep{DBLP:journals/corr/WenGKMSVY15,DBLP:conf/emnlp/WenGMSVY15} and more
recent systems designed for end-to-end training~\citep{DBLP:journals/corr/WenGMRSUVY16}.

\begin{figure}[!htb]
    \centering
    \includegraphics[width=0.7\textwidth]{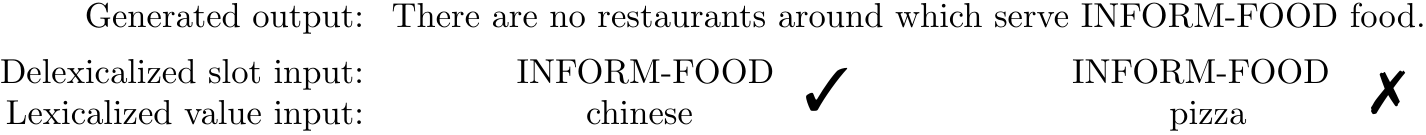}
    \vspace*{-3mm}
    \caption{Models which use only delexicalized slots as input often generate
             grammatically incorrect sentences, since the correct grammatical form
             depends on the lexicalized slot-values.}
    \label{fig:lexadvantage}
    \vspace*{-3mm}
\end{figure}

In the context of task-oriented dialog systems, the NLG task consists of translating
one or multiple dialog act slot-value pairs,~i.e. (INFORM-NAME, \emph{Super Ramen}),
(INFORM-AREA, \emph{near the plaza}) into a well-formed
sentence~\citep{rajkumar-white-espinosa:2009:NAACLHLT09-Short}
such as ``Super Ramen is located near the plaza''.
Existing RNN-based models~\citep{DBLP:journals/corr/WenGKMSVY15} tackle this problem by relying only on the
\emph{delexicalized} part of the act slot-value pairs,~i.e. the model only considers the act and slot (e.g. INFORM-NAME) and ignores the lexical values (e.g. \emph{Super Ramen}).
\citet{DBLP:conf/emnlp/WenGMSVY15} propose a model that can use lexicalized values. However, since they do not align slots with their values, the model
has no way of knowing which value corresponds to which slot.
These methods ignore linguistic relationships in the lexicalized part of a slot-value pair (e.g. between the words ``near'', ``the'', and ``plaza''),
and between the lexicalized part and its surrounding context (e.g. between ``located'' and ``near'').
As illustrated in Figure~\ref{fig:lexadvantage},
ignoring these often leads to grammatically incorrect sentences.

In this paper, we develop an RNN-based approach which considers both lexicalized and
delexicalized dialogue act slot-value pairs. Our model outperforms existing approaches
measured in both automated (BLEU-4~\citep{DBLP:conf/acl/PapineniRWZ02},
METEOR~\citep{Lavie:2007:MAM:1626355.1626389}, ROUGE~\citep{lin2004rouge}, CIDEr~\citep{DBLP:conf/cvpr/VedantamZP15}) and human evaluation. Moreover, we show that
the performance of our model can be improved further by transferring weights from a
pretrained sentence auto-encoder.

%% file: model.tex
\section{Model}
\label{sec:model}

\begin{wrapfigure}{r}{0.55\linewidth}
	\centering
	\vspace*{-55mm}
	\includegraphics[width=\linewidth]{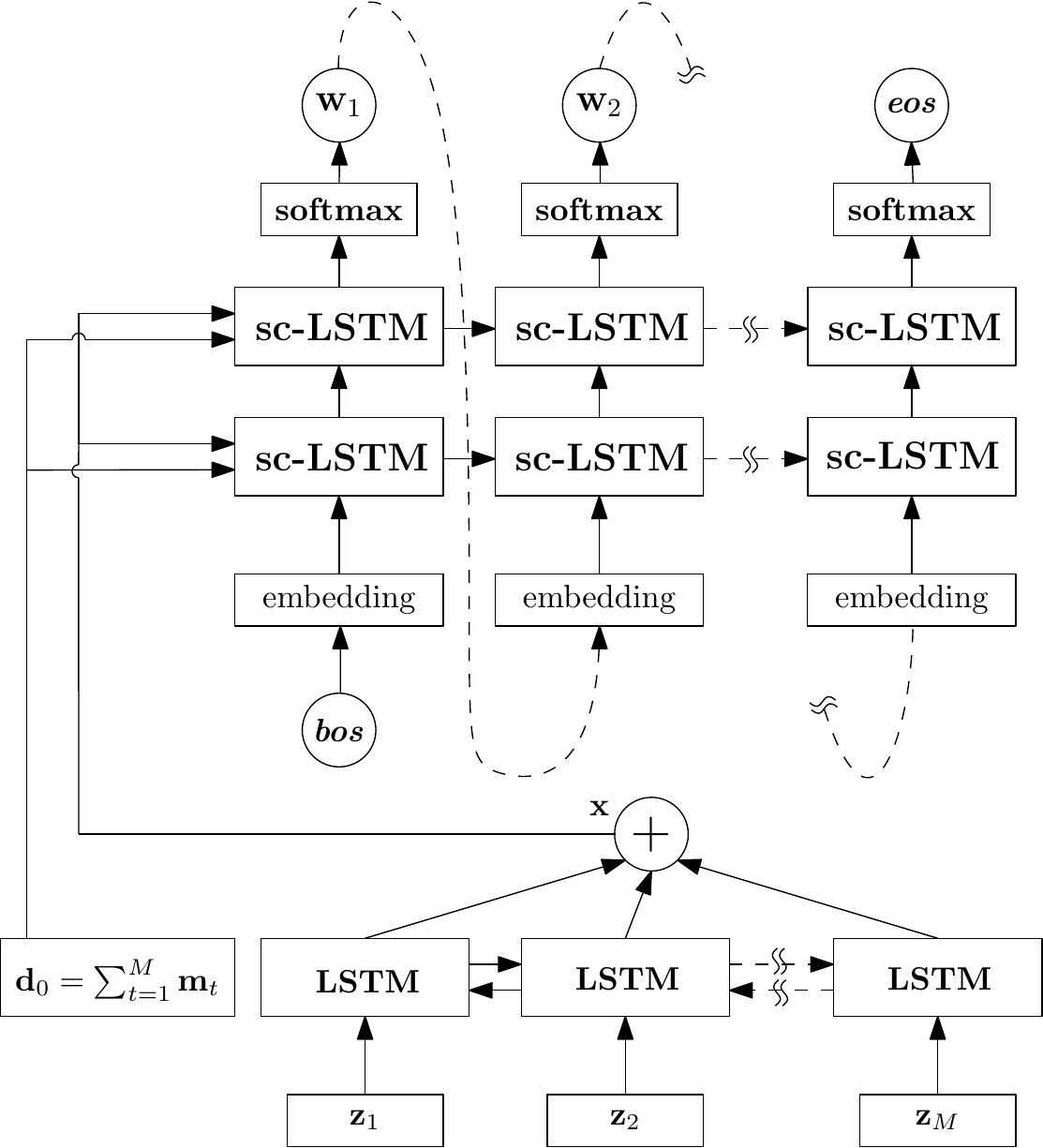}
	\vspace*{-6mm}
	\caption{The encoder-decoder framework for our models: the encoder learns a representation of the
		dialogue act slot-value pairs and the decoder translates them into a natural language sentence.
		\emph{bos} is a special token for the beginning of sentence.
	}
	\label{fig:encoderdecoder}
\end{wrapfigure}

Our model (named ld-sc-LSTM\footnote{lexicalized delexicalized- semantically controlled-
LSTM}) is composed of an RNN encoder and an RNN decoder (see Fig.~\ref{fig:encoderdecoder}).

\begin{wrapfigure}{l}{0.4\linewidth}
    \centering
    \vspace*{-2mm}
    \includegraphics[width=\linewidth]{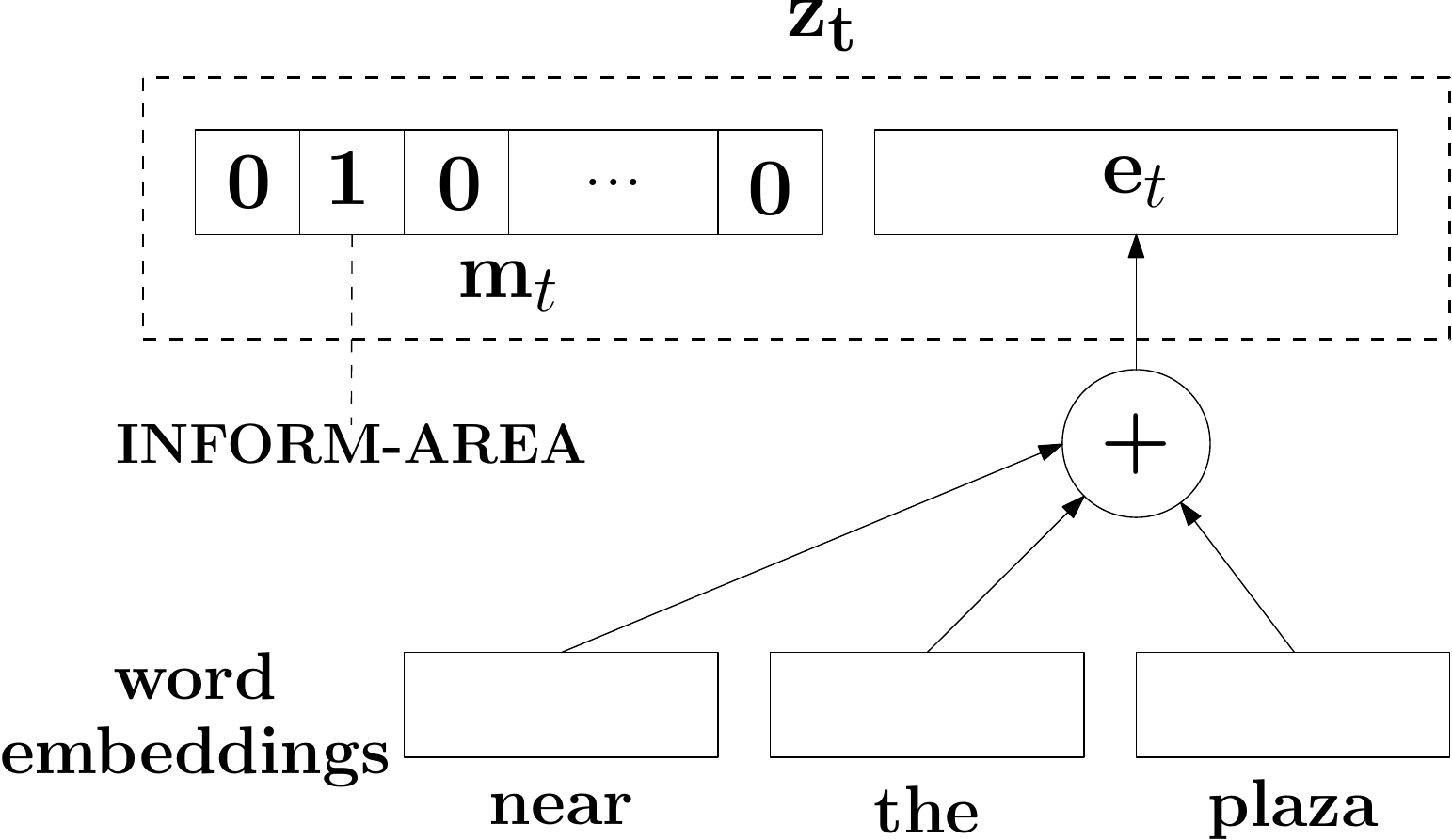}
    \vspace*{-6mm}
    \caption{The encoder input at time-step $t$ for the dialogue act (INFORM-AREA, \emph{near the plaza}).}
    \label{fig:encoderinput}
\end{wrapfigure}

\paragraph{Encoder} The encoder is a 1-layer, bi-directional
LSTM~\citep{DBLP:journals/neco/HochreiterS97}. It takes as input a list of slot-value
pairs for which a sentence must be produced and outputs a representation taking into
account both the delexicalized and the lexicalized parts of each dialogue act slot-value
pair. Particularly, for each input slot-value pair $t$, the encoder receives an
input vector $\mathbf{z}_t$ which is formed by concatenating vectors $\mathbf{m}_t$
and $\mathbf{e}_t$. The vector $\mathbf{m}_t$ is a one-hot encoding of the
delexicalized part. The vector $\mathbf{e}_t$ is formed by taking the mean of the
word embeddings of the lexicalized part. Figure~\ref{fig:encoderinput} illustrates
how the encoder input is created for a given dialogue act.

\paragraph{Decoder} The decoder translates the encoding of the dialogue act slot-value pairs into a fluent sentence.
Our decoder uses sc-LSTM~\citep{DBLP:conf/emnlp/WenGMSVY15} units.
The ``dialogue act'' vector in the sc-LSTM can function similarly to a memory that remembers which acts are yet to be translated~\citep{DBLP:conf/emnlp/WenGMSVY15}.
The initial value of the dialogue act vector is set to $\mathbf{d}_0 = \sum_{t=1}^M \mathbf{m}_t$, where $M$ is the number of encoder time-steps. This is a binary vector whose entries are set to $1$ for the dialogue acts that need to be included in the output sentence.

The encoder information is compressed into a ``context'' vector $\mathbf{x}$ obtained by average-pooling the forward and backward hidden states of the encoder LSTMs across the time-dimension,~i.e. the number of input act slot-value pairs. The vector $\mathbf{x}$ is used to initialize the hidden state $\mathbf{h}_0$ and the memory cell $\mathbf{c}_0$ in the decoder sc-LSTM:
{\small $\mathbf{h}_0 = \tanh(\mathbf{W}_{hx}\mathbf{x}+b_{hx})$,
$\mathbf{c}_0 = \tanh(\mathbf{W}_{cx}\mathbf{x}+b_{cx})$}.
The word embedding of the word output in the previous time-step is also an input to the
decoder sc-LSTM. The hidden states of the sc-LSTM are passed to
softmax layers which produce a word or a delexicalized slot at each time-step. Later,
the slots are replaced with their lexicalized values. The model produces words up to a
predefined maximum length or until it outputs the symbol $eos$.
Our model is summarized in
Figure~\ref{fig:encoderdecoder}.

\paragraph{Loss function and Regularization} We use the negative log-likelihood along with regularization as the loss
function as proposed by \citet{DBLP:conf/emnlp/WenGMSVY15},\\
\begin{equation}
    \small
    \label{eq:loss}
    L = -\sum_{t=1}^T \mathbf{y}_t^\top \log(\mathbf{p}_t) + \left\|\mathbf{d}_T\right\| + \sum_{t=1}^T \eta{\xi^{\left\|\mathbf{d}_{t}-\mathbf{d}_{t-1}\right\|}},\nonumber
\end{equation}
where $\mathbf{y}_t$ is the ground truth word distribution, $\mathbf{p}_t$ is the
predicted word distribution, $T$ is the number of time-steps in the decoder, and $\eta=0.0001$ and
$\xi=100$ are scalars. The term
$\left\|\mathbf{d}_T\right\|$ pushes the model to generate all the slots it is supposed to
generate so that at the last time-step there are no slots remaining. The last term
encourages the model not to drop multiple ``dialogue act'' vector elements at once since the
decoder can only generate one slot/word at each time-step.

\paragraph{Decoding} We use beam search for decoding at test time with a beam width of
10 and slot error rate (ERR) as used in the recent literature \citep{DBLP:conf/emnlp/WenGMSVY15}. The $\lambda$ parameter of the ERR cost was set to $1000$
to severely discourage the decoding from generating sentences which either contain missing or redundant slots.

\paragraph{Transfer learning}
In order to improve the grammar in generated sentences in domains where training data is
limited, we pretrain a sentence auto-encoder on sentences about the same topic, e.g.,
restaurant reviews for our case. The model learns a representation of an input sentence
(with an encoder) and then uses the representation to re-generate the input sentence
(with a separate decoder). The encoder here receives just the word embeddings for the
input sentence (as there are no dialogue acts). The decoder uses LSTM units instead
of sc-LSTM. After training, we use the hidden LSTM weights of the auto-encoder decoder
as initial values of the corresponding hidden weights of the sc-LSTM decoder
($\mathbf{W}_{f}$, $\mathbf{W}_{i}$, $\mathbf{W}_{o}$ and $\mathbf{W}_{c}$) in the
ld-sc-LSTM model. These weights are fine-tuned along with the other weights of the
ld-sc-LSTM model. We name this model transfer-ld-sc-LSTM.

%% file: evaluation.tex
\begin{table*}[!t]
	\small
    \caption{\label{table:cf3-dstc2-metrics} Comparison of performance on the CF and DSTC2 (see Appendix \ref{sec:experiments}) datasets. $^\dagger$ denotes statistical significance (Welch's t-test $p < 0.05$ respectively) with respect to the baselines for \textbf{H}.}
	\centering
	\begin{tabu}to\columnwidth{@{}X[l,4]*5{X[c]}X[c,0.1]*5{X[c]}@{}}
		\toprule
		& \multicolumn{4}{c}{\bf CF}                    & & \multicolumn{4}{c}{\bf DSTC2}                 \\
		\cmidrule{2-6} \cmidrule{8-12}
		\bf Model            & \bf B-4   & \bf M     & \bf R\_L  & \bf C     & \bf H     & & \bf B-4   & \bf M     & \bf R\_L  & \bf C     & \bf H\\
		\midrule
		LSTM                 & 0.277     & 0.284     & 0.502     & 2.080     & 3.552     & & 0.797     & 0.532     & 0.847     & 7.375     & 2.962 \\
		d-sc-LSTM              & 0.291     & 0.288     & 0.513     & 2.231     & 3.504     & & 0.805     & 0.555     & 0.867     & 7.605     & 3.218 \\
		\cmidrule(r){1-1}
		ld-sc LSTM           & 0.308     & 0.293     & 0.518     & 2.329     & \bf 3.838 & & 0.822     & 0.565     & 0.892     & 8.133     & 3.700$^{\dagger}$ \\
		transfer-ld-sc LSTM  & \bf 0.317 & \bf 0.298 & \bf 0.526 & \bf 2.370 & 3.614     & & \bf 0.832 & \bf 0.578 & \bf 0.894 & \bf 8.248 & \bf 3.926$^{\dagger}$ \\
		\bottomrule
	\end{tabu}
	Metrics: BLEU-4 (B-4), METEOR (M), ROUGE\_L (R\_L), CIDEr (C), and Human evaluation (H)
\end{table*}

%% file: results.tex
\section{Results}
\label{sec:results}

We evaluate BLEU-4, METEOR, ROUGE\_L, CIDEr scores using the generated sentence as the
candidate caption and the ground truth as the reference caption. We use publicly
available coco-caption\footnote{\url{https://github.com/tylin/coco-caption}} code to
calculate these metrics. Results for CF and DSTC2 (see Appendix~\ref{sec:experiments}) datasets are shown in
Table~\ref{table:cf3-dstc2-metrics}.
The baseline LSTM, d-sc-LSTM models do not contain our recurrent multi-step lexicalized
encoder. The d-sc-LSTM model differs from the sc-LSTM model of \citet{DBLP:conf/emnlp/WenGMSVY15} in
that it does not use lexicalized values in the dialogue act vector and it does not have a backward reranker.
The ld-sc-LSTM and the transfer-ld-sc-LSTM consistently
perform better than the other baselines in terms of
automated metrics.
All four models use only a forward reranker.

We present the average scores assigned to each model's sentences by five human judges in
Table~\ref{table:cf3-dstc2-metrics}. Our models statistically outperform the baselines for
both datasets. Although transfer-ld-sc-LSTM performs better than ld-sc-LSTM on the DSTC2
dataset, the difference in the scores is not statistically significant. We note that this
may also be due to the limited number of examples in our human rated set which places
constraints on the practicality of p-values. Differences in human evaluation and automated
metrics are to be expected as these metrics do not strongly correlate with human
scores \citep{DBLP:journals/corr/LiuLSNCP16}.

\begin{wraptable}{l}{0.53\columnwidth}
	\small
	\vspace{-7mm}
	\caption{\label{table:cf3-compare} Comparison of top responses generated for some
		dialogue acts on the CF  and DSTC2 datasets.}
	\begin{tabular}{l} \toprule
		\bf (OFFER-NAME, Super Ramen)\\
		\bf (INFORM-FOOD, pizza)\\
		1. Super Ramen serves pizza food.\\
		2. Super Ramen serves pizza food.\\
		3. Super Ramen serves pizza.\\
		4. Super Ramen serves pizza.\\\midrule

		\bf (INFORM-FOOD, pizza)\\
		\bf (INFORM-ADDR, near 108 Queen Street)\\
		1. I am searching for pizza places at near 108 Queen\\
		\quad Street.\\
		2. I am searching for pizza restaurants at near 108\\
		\quad Queen Street.\\
		3. I am searching for pizza places near 108 Queen Street\\
		4. I am searching for pizza places near 108 Queen Street\\\midrule

		\bf (EXPLICIT\_CONFIRMATION-FOOD, dontcare)\\
		1. You are looking for a dontcare restaurant right?\\
		2. You are looking for a dontcare restaurant right?\\
		3. You are looking for a restaurant serving any kind of\\
		\quad food right?\\
		4. You are looking for a restaurant serving any kind of\\
		\quad food right?\\\midrule

		\bf (CANTHELP-FOOD, Japanese) \\
		\bf (CANTHELP-PRICERANGE, under 30 dollars)\\
		1. No Japanese under 30 dollars\\
		2. No Japanese under 30 dollars\\
		3. I'm sorry but there is no Japanese restaurant for under\\
		\quad 30 dollars.\\
		4. There are no Japanese restaurants in under 30 dollars.\\
		\bottomrule
	\end{tabular}
	{\small 1$\rightarrow$ LSTM \hfill 2$\rightarrow$ d-sc-LSTM\\
		3$\rightarrow$ ld-sc-LSTM \hfill 4$\rightarrow$transfer-ld-sc-LSTM}
	\vspace*{-9mm}
\end{wraptable}

Table~\ref{table:cf3-compare} compares responses generated by several models for the
same dialogue acts. In the first example, the LSTM and the d-sc-LSTM generate ``OFFER-NAME serves INFORM-FOOD food.''
since this works with many cuisine values such as Chinese, Indian and Japanese.
In the same example, the ld-sc-LSTM and transfer-ld-sc-LSTM generate ``OFFER-NAME serves INFORM-FOOD.''.
By learning from the lexicalized values of the slots, these models can capture that ``pizza'' should not be followed by ``food''. The remaining examples
also demonstrate our approach's grammatical continuity around generated slots.
Our models do not require any modification to work with special or negative values
like \textit{dontcare}. Overall, we found that
the ld-sc-LSTM and the transfer-ld-sc-LSTM models are less prone to making
grammatical errors, which is also confirmed by their human assessment scores.

%% file: conclusion.tex
\section{Conclusion}
\label{sec:conclusion}

We proposed a recurrent encoder-decoder model for NLG that learns from both lexicalized
and delexicalized tokens. We evaluated our model with several popular metrics used in
the NLP and MT literatures, and also asked humans to evaluate the generated responses.
Our models consistently outperformed existing RNN-based approaches on the CF restaurant
domain dataset and the publicly available DSTC2 dataset. Our transfer learning
experiments showed that bootstrapping with weights from a pretrained sentence auto-encoder can
result in the generation of better responses. Exposing the deep
neural network to the complete data (lexicalized and delexicalized; slots aligned with values) led to a more
powerful model.

%% file: dataset.tex
\label{sec:data}

\paragraph{LMD: restaurant reviews}
This dataset comprises sentences collected from online restaurant reviews. We collected
reviews written in English and sorted them on the basis of highest occurrence of the
words
\textit{phone, postcode, price, food, area, restaurant, nice, address, reservation,}
and \textit{book}. We then trained a sentence auto-encoder on the top $5\,000$ sentences and
used it as a source of pretrained weights for our transfer-ld-sc-LSTM model.

\paragraph{CF: CrowdFlower restaurant search}
We collected this dataset by releasing separate tasks for each dialogue act on
CrowdFlower\footnote{\url{https://www.crowdflower.com}}. The dialogue acts were
\textit{inform, offer, request, implicit confirmation, explicit confirmation, canthelp}.
These dialogue acts were associated with the slots \textit{name, address, phone, area,
postcode, food, pricerange}.
We report results on a test set obtained using a stratified
85\%/15\% train/test split. The dialogue act slot-value pairs were tagged by human
experts after collecting the raw data.

\paragraph{DSTC2: Dialogue State Tracking Challenge 2}
This dataset was extracted from the DSTC2 \citep{henderson2014second} dataset, which
already contains templated machine responses annotated with dialogue acts and slot-value
pairs. The dialogue acts used were
\textit{inform, offer, request, implicit confirmation, explicit confirmation, canthelp,
select, welcome message, repeat, reqmore}, with the same slot types as the CF dataset.

In both CF and DSTC2, the \textit{request} act was allowed to have only empty-valued
slots and for other acts \textit{dontcare} values were allowed in addition to words
from the general vocabulary. We use 10\% of the training set for validation.
Statistics for both datasets are provided in Table~\ref{table:dataset-stats}.

%% file: iclr2017_workshop.bbl
\begin{thebibliography}{22}
\providecommand{\natexlab}[1]{#1}
\providecommand{\url}[1]{\texttt{#1}}
\expandafter\ifx\csname urlstyle\endcsname\relax
  \providecommand{\doi}[1]{doi: #1}\else
  \providecommand{\doi}{doi: \begingroup \urlstyle{rm}\Url}\fi

\bibitem[Bahdanau et~al.(2015)Bahdanau, Cho, and
  Bengio]{DBLP:journals/corr/BahdanauCB14}
Dzmitry Bahdanau, Kyunghyun Cho, and Yoshua Bengio.
\newblock Neural machine translation by jointly learning to align and
  translate.
\newblock In \emph{ICLR}, 2015.

\bibitem[Henderson et~al.(2014)Henderson, Thomson, and
  Williams]{henderson2014second}
Matthew Henderson, Blaise Thomson, and Jason Williams.
\newblock The second dialog state tracking challenge.
\newblock In \emph{SIGDIAL}, volume 263, 2014.

\bibitem[Hochreiter \& Schmidhuber(1997)Hochreiter and
  Schmidhuber]{DBLP:journals/neco/HochreiterS97}
Sepp Hochreiter and J{\"{u}}rgen Schmidhuber.
\newblock Long short-term memory.
\newblock \emph{Neural Computation}, 9\penalty0 (8):\penalty0 1735--1780, 1997.

\bibitem[Karpathy \& Li(2015)Karpathy and Li]{DBLP:conf/cvpr/KarpathyL15}
Andrej Karpathy and Fei{-}Fei Li.
\newblock Deep visual-semantic alignments for generating image descriptions.
\newblock In \emph{{CVPR}}, pp.\  3128--3137, 2015.

\bibitem[Kingma \& Ba(2015)Kingma and Ba]{DBLP:journals/corr/KingmaB14}
Diederik~P. Kingma and Jimmy Ba.
\newblock Adam: {A} method for stochastic optimization.
\newblock In \emph{ICLR}, 2015.

\bibitem[Kondadadi et~al.(2013)Kondadadi, Howald, and
  Schilder]{DBLP:conf/acl/KondadadiHS13}
Ravi Kondadadi, Blake Howald, and Frank Schilder.
\newblock A statistical {NLG} framework for aggregated planning and
  realization.
\newblock In \emph{{ACL}}, pp.\  1406--1415, 2013.

\bibitem[Lavie \& Agarwal(2007)Lavie and
  Agarwal]{Lavie:2007:MAM:1626355.1626389}
Alon Lavie and Abhaya Agarwal.
\newblock Meteor: An automatic metric for {MT} evaluation with high levels of
  correlation with human judgments.
\newblock In \emph{SMT Workshop}, StatMT '07, pp.\  228--231. ACL, 2007.

\bibitem[Lin(2004)]{lin2004rouge}
Chin-Yew Lin.
\newblock Rouge: A package for automatic evaluation of summaries.
\newblock In \emph{Text summarization branches out: ACL-04 workshop}, volume~8,
  2004.

\bibitem[Liu et~al.(2016)Liu, Lowe, Serban, Noseworthy, Charlin, and
  Pineau]{DBLP:journals/corr/LiuLSNCP16}
Chia{-}Wei Liu, Ryan Lowe, Iulian~Vlad Serban, Michael Noseworthy, Laurent
  Charlin, and Joelle Pineau.
\newblock How {NOT} to evaluate your dialogue system: An empirical study of
  unsupervised evaluation metrics for dialogue response generation.
\newblock In \emph{EMNLP}, 2016.

\bibitem[Oh \& Rudnicky(2002)Oh and Rudnicky]{DBLP:journals/csl/OhR02}
Alice Oh and Alexander~I. Rudnicky.
\newblock Stochastic natural language generation for spoken dialog systems.
\newblock \emph{Computer Speech {\&} Language}, 16\penalty0 (3-4):\penalty0
  387--407, 2002.

\bibitem[Papineni et~al.(2002)Papineni, Roukos, Ward, and
  Zhu]{DBLP:conf/acl/PapineniRWZ02}
Kishore Papineni, Salim Roukos, Todd Ward, and Wei{-}Jing Zhu.
\newblock Bleu: a method for automatic evaluation of machine translation.
\newblock In \emph{{ACL}}, pp.\  311--318, 2002.

\bibitem[Rajkumar et~al.(2009)Rajkumar, White, and
  Espinosa]{rajkumar-white-espinosa:2009:NAACLHLT09-Short}
Rajakrishnan Rajkumar, Michael White, and Dominic Espinosa.
\newblock Exploiting named entity classes in ccg surface realization.
\newblock In \emph{HLT-NAACL}, pp.\  161--164. ACL, 2009.

\bibitem[Rieser \& Lemon(2010)Rieser and Lemon]{DBLP:conf/eacl/RieserL10}
Verena Rieser and Oliver Lemon.
\newblock Natural language generation as planning under uncertainty for spoken
  dialogue systems.
\newblock In \emph{Empirical Methods in Natural Language Generation:
  Data-oriented Methods and Empirical Evaluation}, pp.\  105--120, 2010.

\bibitem[Sharma(2016)]{thesis/Sharma16}
Shikhar Sharma.
\newblock Action recognition and video description using visual attention.
\newblock Master's thesis, University of Toronto, 2016.

\bibitem[Srivastava et~al.(2014)Srivastava, Hinton, Krizhevsky, Sutskever, and
  Salakhutdinov]{DBLP:journals/jmlr/SrivastavaHKSS14}
Nitish Srivastava, Geoffrey~E. Hinton, Alex Krizhevsky, Ilya Sutskever, and
  Ruslan Salakhutdinov.
\newblock Dropout: a simple way to prevent neural networks from overfitting.
\newblock \emph{JMLR}, 15\penalty0 (1):\penalty0 1929--1958, 2014.

\bibitem[Vedantam et~al.(2015)Vedantam, Zitnick, and
  Parikh]{DBLP:conf/cvpr/VedantamZP15}
Ramakrishna Vedantam, C.~Lawrence Zitnick, and Devi Parikh.
\newblock Cider: Consensus-based image description evaluation.
\newblock In \emph{{CVPR}}, pp.\  4566--4575, 2015.

\bibitem[Walker et~al.(2007)Walker, Stent, Mairesse, and
  Prasad]{DBLP:journals/jair/WalkerSMP07}
Marilyn~A. Walker, Amanda Stent, Fran{\c{c}}ois Mairesse, and Rashmi Prasad.
\newblock Individual and domain adaptation in sentence planning for dialogue.
\newblock \emph{J. Artif. Intell. Res. {(JAIR)}}, 30:\penalty0 413--456, 2007.

\bibitem[Wen et~al.(2015{\natexlab{a}})Wen, Gasic, Kim, Mrksic, Su, Vandyke,
  and Young]{DBLP:journals/corr/WenGKMSVY15}
Tsung{-}Hsien Wen, Milica Gasic, Dongho Kim, Nikola Mrksic, Pei{-}hao Su, David
  Vandyke, and Steve~J. Young.
\newblock Stochastic language generation in dialogue using recurrent neural
  networks with convolutional sentence reranking.
\newblock In \emph{SIGDIAL}, 2015{\natexlab{a}}.

\bibitem[Wen et~al.(2015{\natexlab{b}})Wen, Gasic, Mrksic, Su, Vandyke, and
  Young]{DBLP:conf/emnlp/WenGMSVY15}
Tsung{-}Hsien Wen, Milica Gasic, Nikola Mrksic, Pei{-}hao Su, David Vandyke,
  and Steve~J. Young.
\newblock Semantically conditioned {LSTM}-based natural language generation for
  spoken dialogue systems.
\newblock In \emph{{EMNLP}}, pp.\  1711--1721, 2015{\natexlab{b}}.

\bibitem[Wen et~al.(2017)Wen, Gasic, Mrksic, Rojas{-}Barahona, Su, Ultes,
  Vandyke, and Young]{DBLP:journals/corr/WenGMRSUVY16}
Tsung{-}Hsien Wen, Milica Gasic, Nikola Mrksic, Lina~Maria Rojas{-}Barahona,
  Pei{-}Hao Su, Stefan Ultes, David Vandyke, and Steve~J. Young.
\newblock A network-based end-to-end trainable task-oriented dialogue system.
\newblock In \emph{EACL}, 2017.

\bibitem[Xu et~al.(2015)Xu, Ba, Kiros, Cho, Courville, Salakhutdinov, Zemel,
  and Bengio]{DBLP:conf/icml/XuBKCCSZB15}
Kelvin Xu, Jimmy Ba, Ryan Kiros, Kyunghyun Cho, Aaron~C. Courville, Ruslan
  Salakhutdinov, Richard~S. Zemel, and Yoshua Bengio.
\newblock Show, attend and tell: Neural image caption generation with visual
  attention.
\newblock In \emph{{ICML}}, pp.\  2048--2057, 2015.

\bibitem[Yao et~al.(2015)Yao, Torabi, Cho, Ballas, Pal, Larochelle, and
  Courville]{DBLP:conf/iccv/YaoTCBPLC15}
Li~Yao, Atousa Torabi, Kyunghyun Cho, Nicolas Ballas, Christopher~J. Pal, Hugo
  Larochelle, and Aaron~C. Courville.
\newblock Describing videos by exploiting temporal structure.
\newblock In \emph{{ICCV}}, pp.\  4507--4515, 2015.

\end{thebibliography}
